\documentclass{article}

\usepackage{arxiv}

\usepackage[utf8]{inputenc} % allow utf-8 input
\usepackage[T1]{fontenc}    % use 8-bit T1 fonts
\usepackage{hyperref}       % hyperlinks
\usepackage{url}            % simple URL typesetting
\usepackage{booktabs}       % professional-quality tables
\usepackage{amsfonts}       % blackboard math symbols
\usepackage{nicefrac}       % compact symbols for 1/2, etc.
\usepackage{microtype}      % microtypography
\usepackage{lipsum}
\usepackage{graphicx}
\usepackage{multirow}
\graphicspath{ {./images/} }

\title{Challenges and approaches to time-series forecasting in data center telemetry: A Survey}

\author{
 Shruti Jadon \\
  Juniper Networks Inc \\
  Sunnyvale, CA \\
  \texttt{sjadon@juniper.net} \\
   \And
 Jan Kanty Milczek  \\
  deepsense.ai\\
  Sunnyvale, CA\\
  \texttt{jan.milczek@deepsense.ai} \\
  \And
 Ajit Patankar \\
Juniper Networks Inc \\
  Sunnyvale, CA \\
  \texttt{apatankar@juniper.net} \\
}

\begin{document}
\maketitle
\begin{abstract}
Time-series forecasting has been an important research domain for so many years. Its applications include ECG predictions, sales forecasting, weather conditions, even COVID-19 spread predictions. These applications have motivated many researchers to figure out an optimal forecasting approach, but the modeling approach also changes as the application domain changes. This work has focused on reviewing different forecasting approaches for telemetry data predictions collected at data centers. Forecasting of telemetry data is a critical feature of network and data center management products. However, there are multiple options of forecasting approaches that range from a simple linear statistical model to high capacity deep learning architectures. In this paper, we attempted to summarize and evaluate the performance of well known time series forecasting techniques. We hope that this evaluation provides a comprehensive summary to innovate in forecasting approaches for telemetry data.
\end{abstract}

\section{Introduction}
Network and data center products collect a large volume of telemetry data such as traffic, CPU utilization, memory usage, etc. One of the most critical business requirements for these products is forecasting the telemetry data.  In broad, forecasting techniques range from simple mean-variance statistical-based methods to deep learning. On top, these look like simply model based choices, but theoretically speaking, these approaches vary in the underlying mathematical formulation, data requirements, programming efforts, and performance as well.

Our work summarizes and evaluates some well representative techniques and demonstrates why one single approach can't fully support telemetry data forecasting.  In this context, our contributions are as follows:

\begin{figure}[ht!]
\centering
  \includegraphics[scale = 0.20]{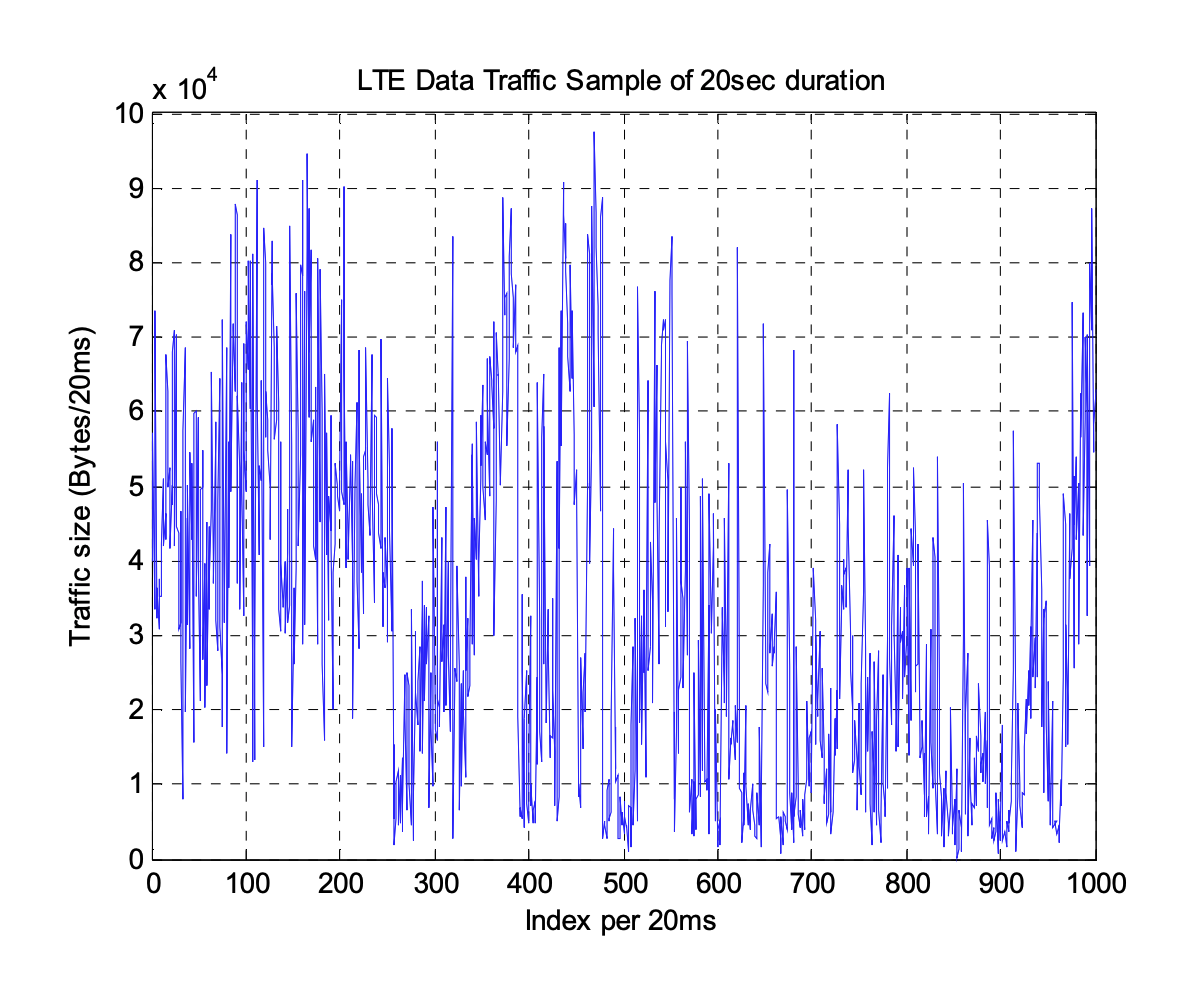}
  \caption{Sample Time-series LTE traffic data for 20sec duration at 20msec index \cite{polaganga2015self}.}
\end{figure}

\begin{enumerate}
    \item Collection and exploratory analysis of telemetry data sets. 
    \item A unified arbitrary length prediction generator.
    \item Overview of important forecasting models.
    \item Evaluation and benchmarking of forecasting models.
    \item Outline of the proposed proprietary model.
\end{enumerate}

The paper is organized as follows: Section II defines the problem definition and requirements or forecasting models' assumptions. In Section III, we theoretically evaluate some widely used time series based forecasting techniques. Our experimental results are listed in Section IV on several real-world data-sets. We then finally conclude the outcomes in section IV. 

\section{Problem definition and requirements}

Let \(T_o,T_1,..T_k \) be time series instantiations of k factors that determine the response variable \(y_0,y_1,..,y_n \).  Our business problem is two-fold:

Single or Next Period Prediction:

\begin{equation} \label{eq: 1}
 y_{n+1}=f(T_0,T_1, ,T_k;y_0,y_1, \ldots,y_n) 
\end{equation}

Multi Period Prediction:
\begin{equation} \label{eq: 2}
y_{n+1, n+2, \ldots n+m}=f(T_0,T_1, ,T_k;y_0,y_1, \ldots,y_n)
\end{equation}

Note that this formulation is different than forecasting \({n+1}^{th}  \) value for each factor.

\subsection{Multi-Period Forecasting}
There are various approaches to generating forecasts for multiple time periods:
\begin{itemize}
    \item fixed-length forecast. The model outputs a fixed-size forecast based on the training data. The model does not expose its internal state after forecasting. This is very similar in concept to the multiple-point rolling prediction, but implementation details make it hard to use that way.
    \item arbitrary-length forecast. The system models the output as a function of time, so it can predict arbitrarily into the future. This can be interpreted as a version of the unified arbitrary-length forecast that simply forecasts the same values at every bootstrapped sample.
    \item single-point rolling prediction. The model predicts a single data point based on the input data. It exposes the internal state, allowing for updating it with the prediction and generating an arbitrary number of data points. Often uses noise to simulate multiple runs and generate confidence intervals/more accurate predictions.
    \item fixed multiple-point rolling prediction. This is similar to the single-point method but the prediction is performed in batches.
\end{itemize}

\subsection{Challenges in Multi-Period Forecasting}

The fundamental problem with multi-period forecasting is that in equation \ref{eq: 2} the \(y_{n+1} \) value is unknown when generating forecast for the \(y_{n+2} \) time period.  Our exposition treats this problem in a consistent fashion as follows:

\begin{itemize}
    \item Consistent treatment in modeling phase. Some of the modeling techniques, such as LSTM, have a built-in capability for multi-period forecasting. However, the number of future periods has to be specified during the training phase and this is a significant restriction. Thus, in rest of the paper,  we use only the single period forecasting feature of techniques that may inherently support multi-period forecasting.
    \item Consequentially, the regularizing factor of multi-period forecasting\cite{multitasklearning} is lost, but consistency against well-known techniques like ARIMA is kept.
    \item To generate arbitrary-length multi-period forecasting we use an algorithm that is described next.
\end{itemize}

\subsubsection{Unified arbitrary-length prediction generator}
To generate arbitrary-length predictions from single-period predictions, we use the stepwise method of predicting a single datapoint and feeding it back into the prediction model. 

This is coupled with the bootstrapped residuals method\cite{forecastinghyndman}, where forecasts are augmented by samples from the historical residuals to generate multiple ``possible futures''. The final forecast is an ensemble of these scenarios.

This approach has the added benefit of producing confidence intervals which are defined as quantiles of predictions at each time step.

It may noted that some methods either don't need to implement this (e.g. Holt-Winters), or eschew the bootstrapped residuals method (e.g. ARIMA) for speed. For consistency, we view them as if they produced the same forecast at each bootstrapped sample.

\subsection{Functional Requirements}

The main application in a data center environment is for forecasting of high volume of streaming data emanating from network and cloud devices. The challenges posed in this environment are as follows:

\begin{itemize}
    \item Large number of metrics to forecast. The number of metrics in a data center monitoring application may exceed 1000. 
    \item High data volume. Some of the metrics may be sampled at a high frequency resulting in very high data volume.
    \item Minimal or no data for some metrics.
    \item Non-linear and multi-variate interactions among metrics.
    \item Length of forecasting horizon. Forecasting horizon may vary from minutes to months with plausible use cases for each time interval.
    \item Consuming applications that require high performance and low latency.
\end{itemize}

\section{Forecasting Techniques}

At a high level, forecasting methods can be grouped as those explicitly modeling time series patterns such as state space models (SSM) and those without explicit formulations such as deep learning.  Prominent SSM include ARIMA and exponential smoothing. SSMs are particularly well-suited for applications where the structure of the time series is well-understood as they explicitly incorporate structural assumptions in the model. This allows for a model to be interpretable but requires significant efforts to determine the structure and covariates.  Also, the traditional SSMs cannot infer shared patterns from multiple data sets of similar time series as they have to be fitted separately on each time series. This makes creating forecasts for a new time series challenging.

Non-SSM models like Deep Neural Networks make very few, if any, structural assumptions regarding the underlying process and yet can identify complex patterns within and across multiple time series. These networks however do require significantly more data for training and suffer from a lack of interpretability.

Recently, researchers have proposed hybrid methods that combine both SSM and deep learning techniques \cite{rangapuram2018deep} with promising results. However, these techniques are not yet available in standard AI/ML libraries and are not considered in this paper.

\subsection{State Space Models (SSM)}

\subsubsection{ARIMA (autoregressive integrated moving average)}
ARIMA\cite{arima} is perhaps the most widely used time series forecasting model.  It is characterized by three factors: \( ARIMA(p,d,q) \) where p is the order (no of time lags) of the auto-regressive component, d is degree of differencing, q is the order of moving average model.

Pros:
\begin{itemize}
    \item A standard bench marking technique
	\item Fast performance and easy to implement using many popular libraries
	\item Foundation block for numerous enhancements such as SARIMA (seasonal ARIMA)
	\item Works with limited data and limited computing power
\end{itemize}

Cons:
\begin{itemize}
	 \item Theoretical limitations such as the error terms needs to have a random normal distribution.
	 \item Not suitable for distributions with fat tails or volatility clusters.
	 \item An univariate technique
\end{itemize}

Some of multivariate extensions to ARIMA technique include:
\begin{enumerate}
	 \item Set of dependent variables with a regression for each one. These models are referred as VAR (vector autoregressive models) or VARMA’s
	 \item Set of independent variables (exogenous) for a single dependent variable – these fall under ARIMAX models.
\end{enumerate}

However, there do not appear to be many references of successful applications of the multivariate extensions in practice.

\subsection{EWMA (Exponential Weighted Moving Average)}

In the following exposition, we only consider one period forecast as explained earlier. A simple EWMA\cite{forecastinghyndman} process is defined as follows:

\[y_{t+1} = \alpha y_t+(1-\alpha) y_{t-1} \]
with the initial condition of

\[y_0 = y_0 \]

where, \(y_{t+1}\) is the forecast at \(y_t \).

Holt’s extension: Holt’s extension incorporates slope or trend in the EWMA model.

\[l_t= \alpha y_t+(1-\alpha)(l_{t-1}+b_{t-1} ) \]
\[b_t= \beta (b_t-s_{t-1})+(1-\beta) b_{t-1} \]
\[y_{t+1} = l_t +  y_{t-1} \] 

Holt-Winter extension: This is a further extension of Holt’s model with an additional term for seasonality.

\[l_t= \alpha y_t+(1-\alpha)(l_{t-1}+b_{t-1} ) \]
\[b_t = \beta (l_t - l_{t-1}) + (1- \beta)b_{t-1} \]
\[s_t = \gamma (y_t - l_t) + (1 - \gamma)s_{t-1}  \]
\[y_{t+1} = l_t + b_t + s_t \] 

Pros:

\begin{itemize}
	\item Implementation can be high performance with limited memory.  It is possible to implement as an online model.
	\item Simple model with good interpretability.
    \item Natively supports arbitrary length forecasts
\end{itemize}

Cons:

\begin{itemize}
	\item Model may be too simple to represent real life scenarios.
\end{itemize}

\subsection{GARCH (Generalized Autoregressive Conditional Heteroskedastic Model)}

GARCH\cite{garch} is a more powerful model as it supports heteroskedastic processes:

\begin{enumerate}
\item Autoregressive (AR) Forecast for next time period is a regressed function of time series value in the current time period.
\item Conditional (C).  Forecast for next time period is conditional on the value in the current time period.
\item Heteroskedastic (H). Variance of values is not constant over time period.
\end{enumerate}	

Pros:

\begin{itemize}
\item A powerful modeling technique capable of supporting several common time series processes Issues
\end{itemize}	

Cons:

\begin{itemize}
\item Univariate in forecasting nature.
\item Tuning of parameters is complex
\end{itemize}

\subsubsection{Seasonal Trend Decomposition (STD) Predictor}
We have developed an extension of linear regression technique that incorporates seasonal trends. The linear regression component is based on the time series while seasonality is modeled by transforming time and holidays into categorical features. This approach achieves fast decomposition into linear trend and seasonal factor, and is augmented by input sequence transformations to model more complex patterns.

Equation
\[ LR - \textrm{standard linear regression} \]
\[ F_{time}(t) - \textrm{categorical features based on time} \]
\[ \hat{y}_t = LR(t) + LR(F_{time}(t)) \]

Pros:
\begin{itemize}
\item	Very fast and Easy to use.
\item	Supports calculations of confidence intervals
\item	Supports multivariate forecasting 
\item Supports an arbitrary length forecasts
\end{itemize}	
Cons:
\begin{itemize}
\item	Very simple, cannot model complex relations.
\item	Confidence intervals are over-simplified
\end{itemize}	

\subsubsection{Seasonal Trend Autoregressive (STAR) Predictor}
This predictor extends the STD model by incorporating an autoregressive component. The formulation is as follows:

\[ LR - \textrm{standard linear regression} \]
\[ F_{time}(t) - \textrm{categorical features based on time} \]
\[ aw - \textrm{autoregression window} \]
\[ \hat{y}_t = LR(t) + LR(F_{time}(t)) + LR(y_{t-aw\,:\,t}) \]

Pros:
\begin{itemize}
\item	Fast and Easy to use
\item	Supports basic confidence intervals
\item	extensible to multivariate forecasting
\end{itemize}	

Cons:
\begin{itemize}
\item	More complex than STD Predictor, prone to overfitting
\item	Slower than STD Predictor
\end{itemize}	

\subsection{Specialized Libraries}
\subsubsection{FB (Facebook) Prophet}

Facebook Prophet \cite{fbprophetrefpaper} is a specialized time series library and not a specific statistical technique.  The library is based on an additive model where non-linear trends are fit with yearly, weekly, and daily seasonality, plus holiday effects. It works best with time series that have strong seasonal effects and several seasons of historical data. The library internally uses well known \emph{Stan} statistical library.

Pros:
\begin{itemize}
\item Very easy to use and most of the implementation details are abstracted from the user.
\item Extensive utility and plotting support.
\item Univariate forecasting is well supported.
\item A very good default choice for simple applications
\end{itemize}

Cons:
\begin{itemize}
\item For multi-variate forecasting, the library implementation is not clear.  It appears that future values of additional regressors are needed to make forecast at time values \((t,t+1,\ldots,t+n)\). Furthermore, only these future values are used and not their historical values.
\item Library lacks scalability, flexibility, and extensibility to support complex telemetry forecasting applications.
\end{itemize}

\subsubsection{GluonTS}

GluonTS\cite{gluon} is a deep learning library for time series modeling. It is a fairly self contained library that includes most of the components necessary to build, train and run time series models.

The main issue with this library is that it is based on Apache MXNet deep learning library. Apache MXNet support community is much smaller than either Tensorflow or PyTorch. As we do not want to introduce an additional deep learning framework in our production code, it was decided not to evaluate GluonTS.

% GluonTS is a library for deep learning based time series modeling. It is a fairly self contained library that includes most of components necessary to build, train and run time series models.

% GluonTS includes multiple implementations of state-of-the-art time series processing architectures.

Pros:
\begin{itemize}
\item	Powerful API, a wide variety of models to choose from
\item Natively generates arbitrary length forecast
\end{itemize}

Cons:
\begin{itemize}
\item	Slow
\item	Uses MXNet
\item   Created for short-horizon predictions, rolling predictions currently not implemented and have to be custom implemented
\end{itemize}

\subsection{Deep Learning Models}
Sequence based deep learning architectures are able to learn complex time-series features. These models have flexible capacity and along with large volume of data can lead to excellent model performance.

Pros:
\begin{itemize}
\item	Ability to model very complex relations in the data
\item	State of the art in time series processing
\item	Possibility of using transfer learning
\end{itemize}

Cons:
\begin{itemize}
\item	Training is computationally intensive
\item	Hard to tune/use/automate
\end{itemize}
In the following sections, we review relevant variants of deep learning networks that are applicable for time series forecasting.
\subsubsection{Recurrent Neural Networks}
Recurrent Neural Network is one of the first sequential deep learning architecture. As the name suggests, it is a single layer that is stacked multiple times in order to capture the complexity of a series.

Pros:
\begin{itemize}
\item Ability to learn complex features of time-series.
\end{itemize}

Cons:
\begin{itemize}
\item Vanishing gradient problem
\item Fixed size input is required for training.
\end{itemize}

\begin{figure}[ht!]
\centering
  \includegraphics[scale = 0.22]{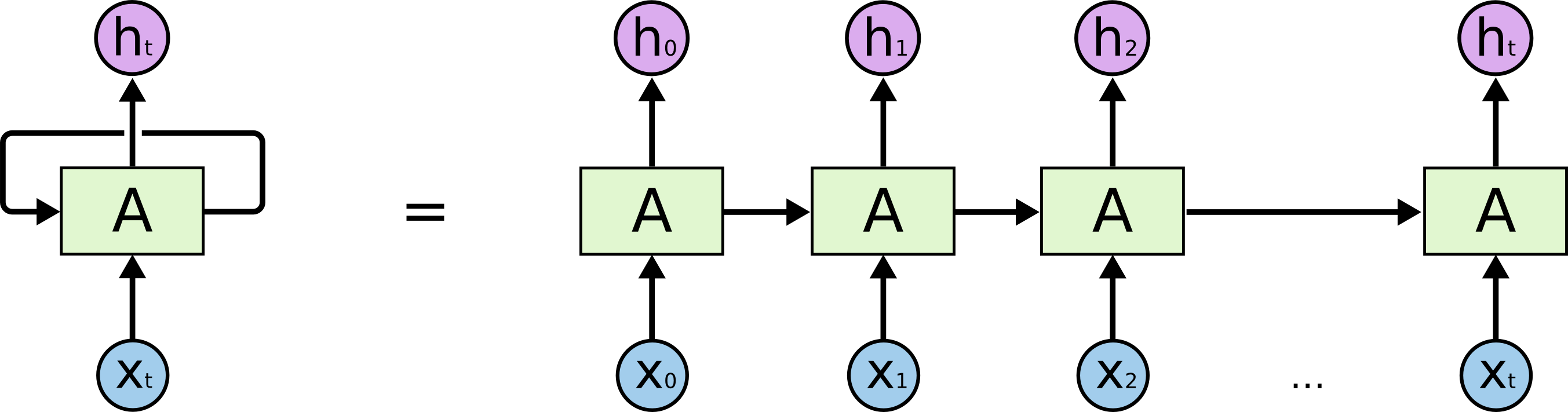}
  \caption{A Visual Representation of Unrolled Recurrent Neural Network\cite{understandinglstmnetworks}}
\end{figure}

\subsubsection{Long Short Term Memory (LSTM) Networks}
LSTMs are a special kind of RNNs, capable of learning long-term dependencies. They work tremendously well on a large variety of problems, and are now widely used.
LSTMs are explicitly designed to avoid the vanishing gradient problem. The architecture inherently remembers information over a long sequence of the time series.

Cons:
\begin{itemize}
\item Computational intensity of training
\item Require large amount of data to train than other models
\item Fixed size input is required for training.
\end{itemize}
\begin{figure}[ht!]
\centering
  \includegraphics[scale = 0.25]{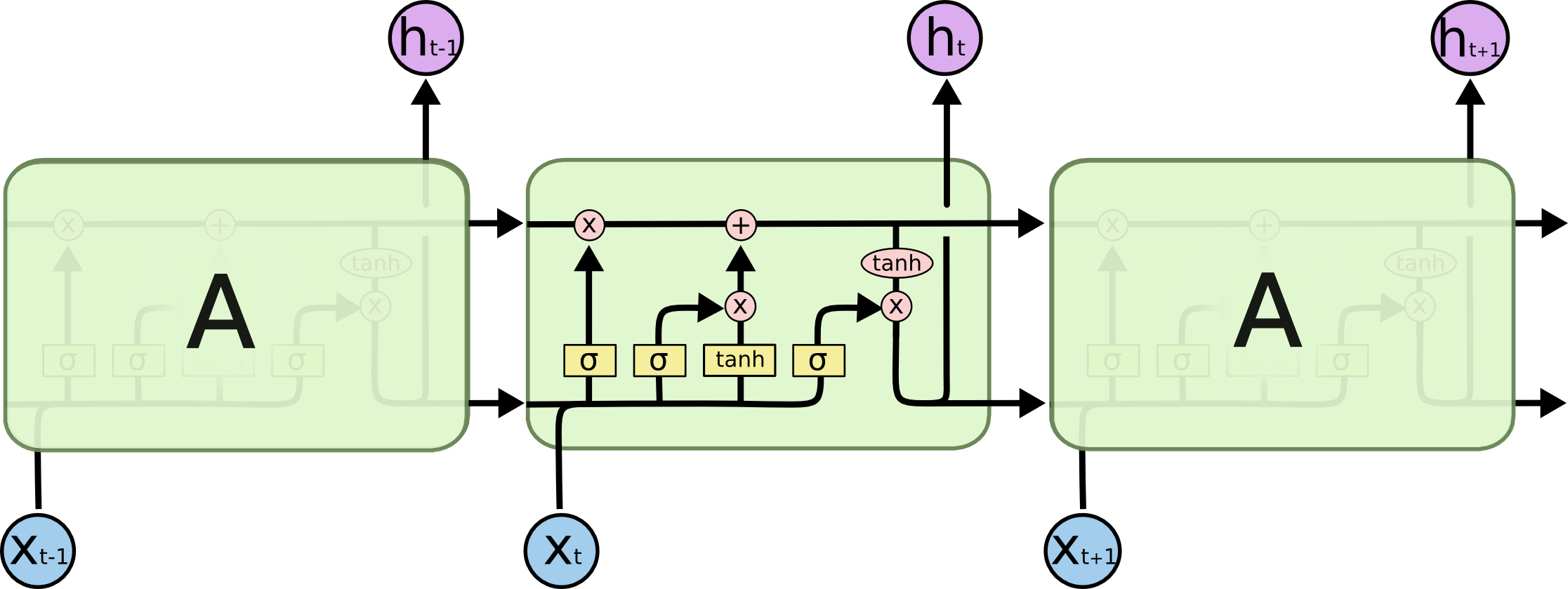}
  \caption{Long Short Term Memory Networks Architecture\cite{understandinglstmnetworks}}
\end{figure}

\subsubsection{Bi-directional LSTMs}
LSTM in its core, preserves information from inputs that has already passed through it using the hidden state. But, Unidirectional LSTM only preserves information of the past because the only inputs it has seen are from the past.
On the other hand bidirectional will run your inputs in two ways, one from past to future and one from future to past and what differs this approach from unidirectional is that in the LSTM that runs backwards you preserve information from the future and using the two hidden states combined you are able in any point in time to preserve information from both past and future.
In general, Bidirectional LSTMs show very good results as they can understand context better. \\
Pros:
\begin{itemize}
    \item Preserves information from future as well as past.
    \item proven to perform better than uni-directional LSTM in learning time-series distribution.
\end{itemize}
Cons:
\begin{itemize}
\item can take a long time to run,
\item require large amount of data to train, and
\item Fixed size input is required for training.
\end{itemize}
\begin{figure}[ht!]
\centering
  \includegraphics[scale = 0.30]{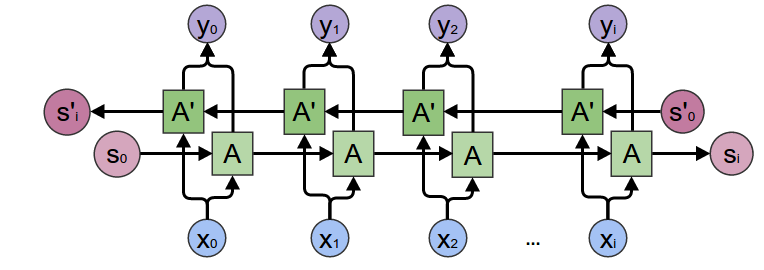}
  \caption{Bi-directional LSTM Architecture\cite{understandinglstmnetworks}}
\end{figure}

\subsection{Probabilistic Models}
Apart from statistical and deep learning models which are majorly parametric approaches, we have also experimented with probabilistic models. In probabilistic models, instead of casting data distribution into another distribution like we do in deep learning approaches, We learn the parameters of it using optimization approaches like Expectation Maximization.
\subsubsection{Hidden Markov Models}
The Hidden Markov Model is based on Markov chain. A Markov chain is a model
that tells us something about the probabilities of sequences of random variables or states. But a Markov chain makes a very strong assumption that for a prediction way too in future, we have dependence on only current state, previous states have no impact on the future except through the current state. So, as part of Hidden Markov Model, we need to calculate all states and transition probability of markov chain, in which sometimes not all states are visible, therefore it's called Hidden Markov Model.A hidden Markov model (HMM) allows us to talk about both observed and hidden events that we think of as causal factors in our probabilistic model. For our case we have taken Gaussian Mixture Models as distribution of states. \\
Pros:
\begin{itemize}
    \item gives better predictions even with less amount of data, and
    \item faster to train and predict.
\end{itemize}
Cons:
\begin{itemize}
    \item Not good for predictions in long run due to its assumption of dependence only on current state.
\end{itemize}
\begin{figure}[ht!]
\centering
  \includegraphics[scale = 0.25]{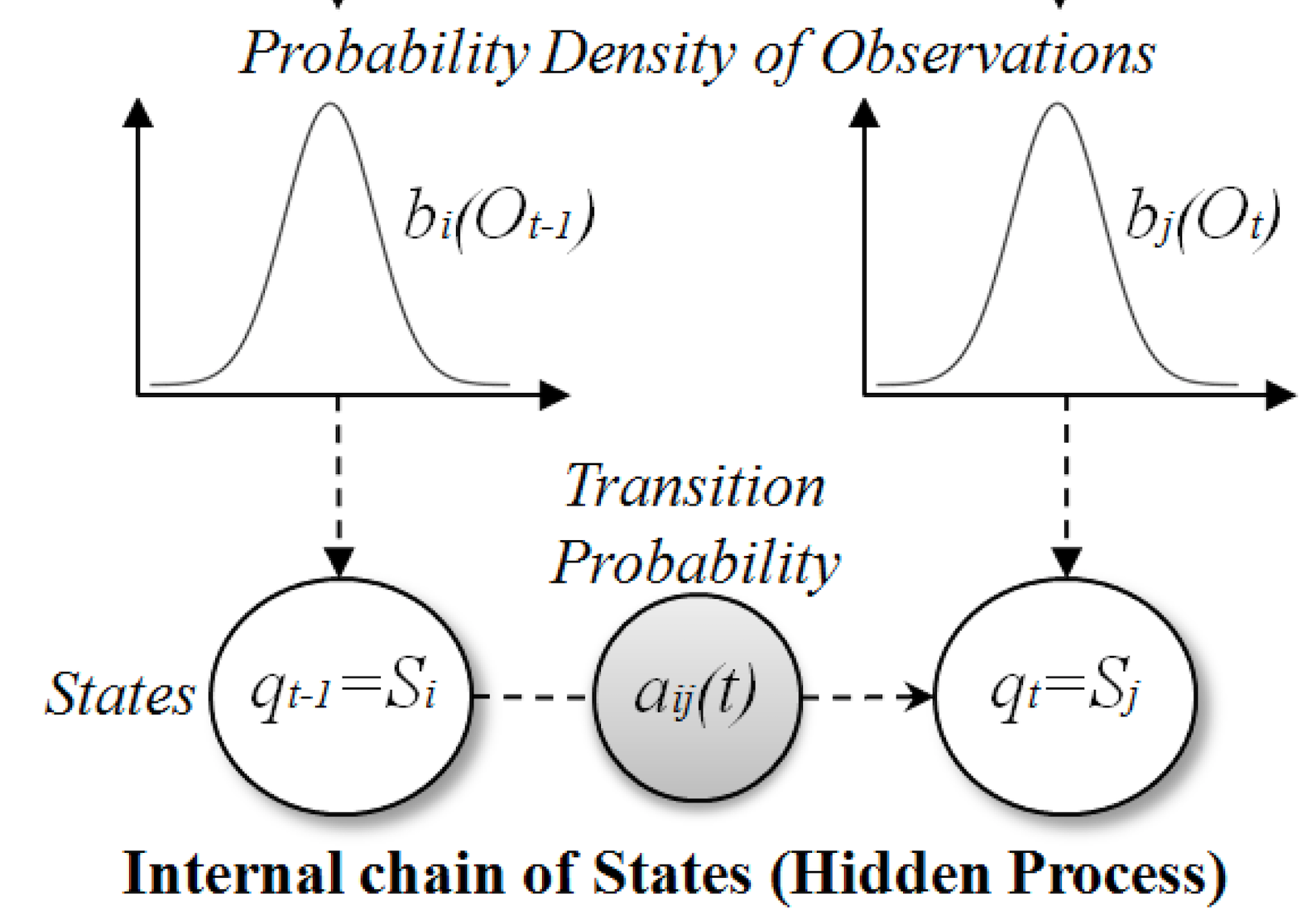}
  \caption{Hidden Markov Model overview\cite{hiddenmarkovgraph}}
\end{figure}

\section{Experiments and Results}

To quantify the differences in runtime and accuracy of the different forecasting techniques, we run them on multiple datasets and record their performance.The datasets were selected to model real-system use cases.
For each dataset, the runtime and two accuracy metrics ($R^2$ and $R^2$ of logarithms) are recorded.

\subsection{Dataset description}
For performance evaluation purpose, we employed four data-sets as listed below: 
\begin{enumerate} 
\item Curtin\cite{curtin} - a dataset of flow data, annotated with unclear timestamps - they were set to simulate a slight hourly seasonality. Set to polling rate of 1min, 10 days total. Has two fields - flow\_size and count.
\item LTE Traffic\cite{ltetraffic} - a dataset of flow data from a mobile carrier's LTE network, hourly, full year
\item Juniper - an internat dataset from Juniper, sampling rate of 15min, over a year in length
\item Twitter\cite{twittervolume} - volume of tweets from 2015, 5min aggregation period, 1.5 months in total
\end{enumerate}

\begin{figure}
\centering
  \includegraphics[scale = 0.70]{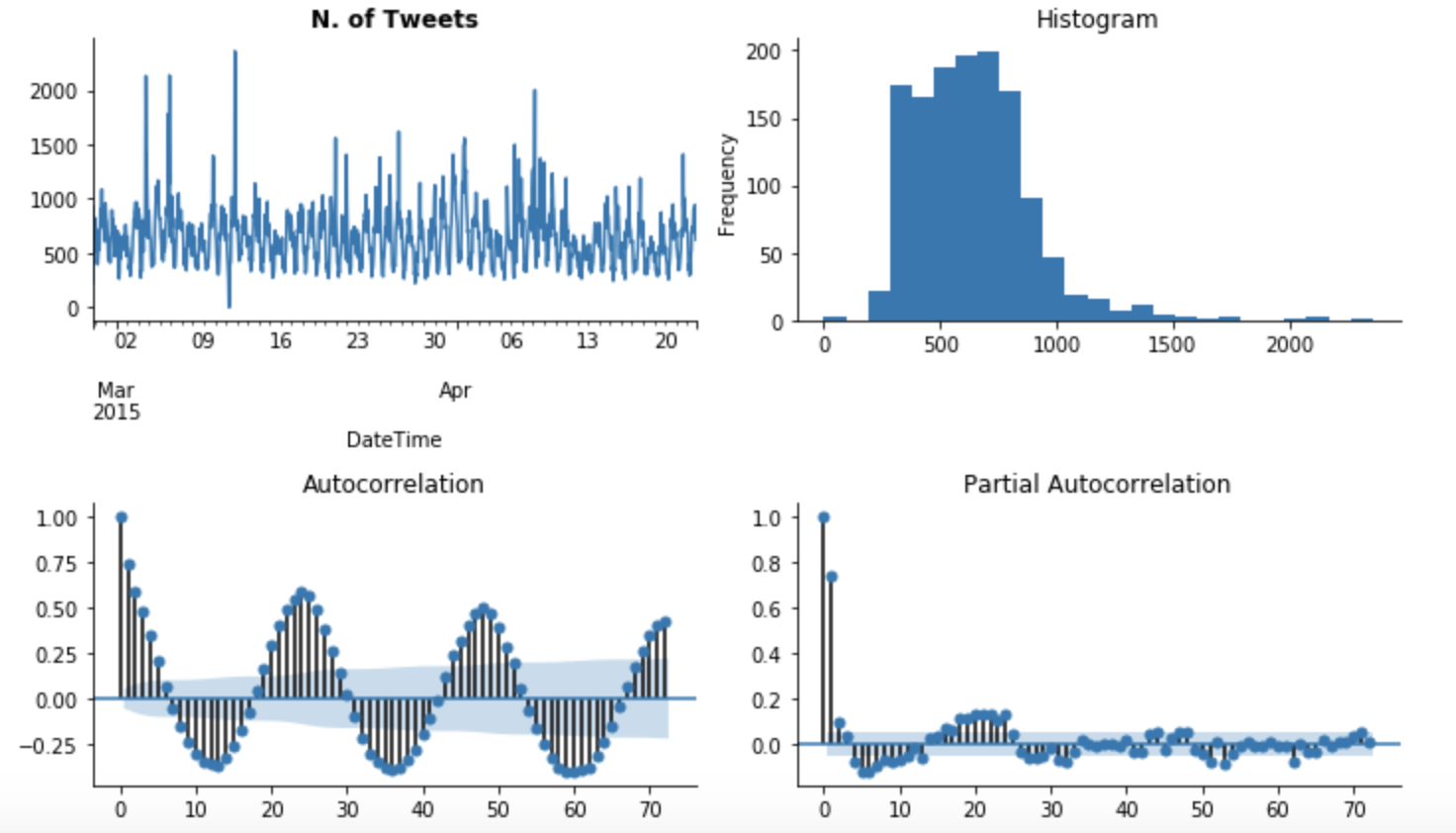}
  \caption{Sample Twitter traffic Dataset\cite{twittervolume}}
\end{figure}

\subsection{Dataset purpose}
\begin{itemize} 
\item Curtin - consists of several "levels" of possible flow size. Simulates a system that receives heavy requests, while otherwise maintaining a constant noise. As the dataset is very hard to fit for (signal to noise ratio is very small), it is mostly used to test for model's resilience to overfitting.
\item LTE Traffic - simulates human behavior on a local (country-wide) scale. Exhibits high daily patterns of sleep-work-free time.
\item Juniper - simulates local human behavior in an exponentially growing system
\item Twitter\cite{twittervolume} - simulates global human behavior with concentration (i.e. users around the world, most from USA).
\end{itemize}
\subsection{Experimental settings}
\textbf{CPU:} Intel(R) Core(TM) i7-6700K CPU @ 4.00GHz\\
\textbf{RAM:} 4x 8GB, 2133 MT/s\\
\textbf{GPU:} GeForce GTX 1060, 6GB

\subsection{Evaluation Metrics}
To assess the performance of each model, we have employed three metrics as listed below:
\begin{itemize}
    \item RT(s) - run time in seconds. This includes data manipulation
    \item $R^2$ - supporting metric
    \item $lR^2$ - $R^2$ on logarithms. Main metric (for traffic, predicting order of magnitude is more important than value)
\end{itemize}

\subsection{Benchmark}
Most models were run multiple times with different parameters, with best results achieved shown here. Each model was benchmarked twice - predicting the last 1000 or 5000 data points with all other data points available for training. One exception was LTE traffic dataset, where 4380 data points (exactly half) were used in place of 5000.

\subsubsection{LTE Traffic}
\centering
\begin{tabular}{|l||c|c|c|c|c|c|}
    \hline
    \multirow{2}{*}{Model} & \multicolumn{3}{|c|}{1000 datapoints} & \multicolumn{3}{|c|}{5000 datapoints} \\
    \cline{2-7}
    & RT(s) & $R^2$ & $lR^2$ & RT(s) & $R^2$ & $lR^2$ \\
    \hline \hline
GRU & 40.4 & 0.77 & 0.73 & 73.12 & 0.69 & 0.6\\ \hline
STAR & 1.27 & \textbf{0.88} & 0.81 & 1.05 & 0.66 & 0.65\\ \hline
FBP & 4.31 & 0.8 & 0.79 & 3.28 & 0.65 & 0.61\\ \hline
HWES & 0.86 & 0.8 & 0.77 & \textbf{0.5} & 0.63 & 0.61\\ \hline
STD & \textbf{0.82} & 0.88 & \textbf{0.82} & 0.6 & \textbf{0.8} & \textbf{0.69}\\ \hline
% RNN & \textbf{0.82} & 0.88 & \textbf{0.82} & 0.6 & \textbf{0.8} & \textbf{0.69}\\ \hline
% LSTM & \textbf{0.82} & 0.88 & \textbf{0.82} & 0.6 & \textbf{0.8} & \textbf{0.69}\\ \hline
% Bi-LSTM & \textbf{0.82} & 0.88 & \textbf{0.82} & 0.6 & \textbf{0.8} & \textbf{0.69}\\ \hline
\end{tabular}

\subsubsection{Curtin - flow size}
\centering
\begin{tabular}{|l||c|c|c|c|c|c|}
    \hline
    \multirow{2}{*}{Model} & \multicolumn{3}{|c|}{1000 datapoints} & \multicolumn{3}{|c|}{5000 datapoints} \\
    \cline{2-7}
    & RT(s) & $R^2$ & $lR^2$ & RT(s) & $R^2$ & $lR^2$ \\
    \hline \hline
GRU & 38.81 & 0.0 & -0.01 & 67.53 & -0.01 & -0.01\\ \hline
STAR & 2.08 & -0.01 & -0.01 & 1.87 & -0.02 & -0.04\\ \hline
FBP & 1.89 & -0.01 & -0.07 & 1.77 & -0.02 & \textbf{0.0}\\ \hline
HWES & 0.92 & -0.01 & \textbf{0.0} & 0.62 & -0.02 & -0.01\\ \hline
STD & 1.34 & -0.01 & -0.01 & 1.09 & -0.02 & -0.02\\ \hline
LSTM & \textbf{0.84} & 0.086 & -1.74 & \textbf{0.34} & 0.03 & -1.7\\ \hline
Deep-LSTM & 1.61 & 0.15 & -1.92 & 0.65 & 0.06 & -1.74\\ \hline
Bi-LSTM & 1.45 & \textbf{0.21} & -1.39 & 0.47 & \textbf{0.09} & -1.03\\ \hline

\end{tabular}

\subsubsection{Curtin - flow count}
\centering
\begin{tabular}{|l||c|c|c|c|c|c|}
    \hline
    \multirow{2}{*}{Model} & \multicolumn{3}{|c|}{1000 datapoints} & \multicolumn{3}{|c|}{5000 datapoints} \\
    \cline{2-7}
    & RT(s) & $R^2$ & $lR^2$ & RT(s) & $R^2$ & $lR^2$ \\
    \hline \hline
GRU & 29.35 & 0.01 & 0.0 & 80.8 & -0.06 & -0.06\\ \hline
STAR & 2.04 & -0.03 & -0.03 & 1.87 & -0.11 & -0.1\\ \hline
FBP & 2.37 & -0.09 & -0.08 & 2.14 & -0.12 & -0.14\\ \hline
HWES & 4.32 & -0.47 & -0.41 & \textbf{0.08} & -0.16 & -0.2\\ \hline
STD & 1.24 & -0.06 & -0.05 & 1.08 & -0.12 & -0.11\\ \hline
LSTM & \textbf{0.73} & 0.52 & 0.52 & 0.51 & 0.53 & 0.54\\ \hline
Deep-LSTM & 1.29 & 0.52 & 0.52 & 0.66 & \textbf{0.54} & \textbf{0.55}\\ \hline
Bi-LSTM & 1.15 & \textbf{0.53} & \textbf{0.53} & 0.69 & 0.53 & \textbf{0.55}\\ \hline

\end{tabular}

\subsubsection{Juniper data-set}
\centering
\begin{tabular}{|l||c|c|c|c|c|c|}
    \hline
    \multirow{2}{*}{Model} & \multicolumn{3}{|c|}{1000 datapoints} & \multicolumn{3}{|c|}{5000 datapoints} \\
    \cline{2-7}
    & RT(s) & $R^2$ & $lR^2$ & RT(s) & $R^2$ & $lR^2$ \\
    \hline \hline
GRU & 42.74 & 0.82 & 0.87 & 81.42 & 0.78 & \textbf{0.88}\\ \hline
STAR & 5.0 & 0.8 & 0.81 & 5.1 & 0.79 & 0.85\\ \hline
FBP & 10.15 & 0.79 & 0.81 & 7.6 & 0.74 & 0.84\\ \hline
HWES & 12.21 & 0.75 & 0.82 & 11.67 & 0.83 & 0.83\\ \hline
STD & 3.77 & 0.64 & 0.79 & 3.45 & 0.63 & 0.83\\ \hline
RNN & 0.746 & 0.978 & 0.943 & \textbf{0.53} & 0.91 & 0.82\\ \hline
LSTM & 0.274 & \textbf{0.98} & \textbf{0.95} & 0.63 & \textbf{0.92} & 0.83\\ \hline
Bi-LSTM & \textbf{0.114} & 0.978 & 0.93 & 4.8 & 0.91 & 0.84\\ \hline

\end{tabular}

\subsubsection{Twitter data-set}
\centering
\begin{tabular}{|l||c|c|c|c|c|c|}
    \hline
    \multirow{2}{*}{Model} & \multicolumn{3}{|c|}{1000 datapoints} & \multicolumn{3}{|c|}{5000 datapoints} \\
    \cline{2-7}
    & RT(s) & $R^2$ & $lR^2$ & RT(s) & $R^2$ & $lR^2$ \\
    \hline \hline
GRU & 30.17 & 0.45 & 0.53 & 69.0 & 0.26 & 0.46\\ \hline
STAR & 2.41 & 0.37 & 0.49 & 2.07 & 0.23 & 0.44\\ \hline
FBP & 4.72 & 0.27 & 0.42 & 2.89 & 0.23 & 0.36\\ \hline
HWES & 1.0 & -0.61 & -0.53 & 0.75 & -0.19 & -0.12\\ \hline
STD & 1.57 & 0.43 & 0.53 & 1.22 & 0.25 & 0.44\\ \hline
LSTM & \textbf{0.79} & 0.58 & 0.61 & \textbf{0.43} & 0.48 & 0.59\\ \hline
Deep LSTM & 1.44 & 0.59 & \textbf{0.62} & 0.62 & \textbf{0.52} & \textbf{0.61}\\ \hline
Bi-LSTM & 1.18 & \textbf{0.60} & 0.61 & 0.59 & 0.46 & 0.59\\ \hline

\end{tabular}

\subsection{Observations}
\begin{enumerate}
    \item Deep Neural Networks were up to 10x slower during training phase than the rest of the algorithms, but they provided the best accuracy and less inference time most often and, curiously, were very resistant to over fitting on the Curtin dataset.
    \item Holt-Winters Exponential smoothing was the fastest (as expected), but was usually outperformed by STD, which does not fall much behind time-wise
    \item Holt-Winters Exponential smoothing heavily overfitted on Twitter data, which is worrying and indicates that sanity checks would be needed if the algorithm was to be used
    \item Both STAR Predictor and FBProphet tended to be outperformed by other methods, but consistently places in the top, making them excellent choices for a single-solution-fits-all approach
    \item STAR and STD were the fastest algorithms by a significant factor on the largest dataset (Juniper), indicating that they should be considered more favorably the more data is to be processed during training.
\end{enumerate}

\subsection{Benchmark conclusions}
\begin{itemize}
    \item Over multiple runs, there was no clear winner on a particular type of dataset
    \item If the system can afford to spend the computational resources, the best approach is to implement deep neural network models.
    \item If the system needs to perform computations quickly, it's recommended to use either FBProphet (out-of-the-box) or STAR (very configurable)
\end{itemize}

% \subsection{Proprietary Algorithm}

% Our proprietary algorithm is based on building a large number of component models and dynamically generating the ensemble based on the following principles:

% \begin{itemize}
% \item A model to match forecast horizon with component models and data source.
% \item Constrained optimization to determine candidate models that can be trained and/or executed in the given response time.
% \item A learning system that generates the output using candidate models.
% \end{itemize}

% This system is currently under development and it will be described in a subsequent publication.

% \begin{figure}[h]
% \vspace{.3in}
% \centerline{\fbox{This figure intentionally left non-blank}}
% \vspace{.3in}
% \caption{Sample Figure Caption}
% \end{figure}

% \subsubsection{Tables}

% \begin{table}[h]
% \caption{Sample Table Title} \label{sample-table}
% \begin{center}
% \begin{tabular}{ll}
% \textbf{PART}  &\textbf{DESCRIPTION} \\
% \hline \\
% Dendrite         &Input terminal \\
% Axon             &Output terminal \\
% Soma             &Cell body (contains cell nucleus) \\
% \end{tabular}
% \end{center}
% \end{table}

\section{Conclusion}
\begin{flushleft}
Time-series forecasting objective is more complex to learn as it requires modeling approaches to associate the correlation factor with time and simultaneously with past observations. A good forecasting model should be able to observe both trends as well as seasonality in data. In this work, we provided a comprehensive review of 11-time series forecasting techniques for telemetry data.These techniques are widely used in other fields, such as stock market prediction, energy consumption forecasting, weather forecasting, etc. We evaluated the above-listed set of representative forecasting techniques on four different time series data-sets and logged our observations as a summary. We can conclude that a single off-the-shelf method is not likely to meet all the requirements due to time-series data's dynamicity.
\end{flushleft}

{\small
\nocite{*}
\bibliographystyle{unsrt}
\bibliography{main}
}

\end{document}